\documentclass[letterpaper, 10 pt, conference]{ieeeconf}  

\IEEEoverridecommandlockouts                              
\usepackage[utf8]{inputenc}
\usepackage{amsmath} 
\usepackage{amssymb}  
\usepackage{mathrsfs}
\usepackage{graphbox}
\usepackage{subcaption}
\usepackage{float}
\usepackage{multicol, blindtext}
\usepackage{color}
\usepackage{soul}
\usepackage{graphicx}  
\usepackage{balance}
\usepackage{dsfont}
\usepackage{hyperref}
\title{\LARGE \bf
Fast Adaptation with Meta-Reinforcement Learning for Trust Modelling in Human-Robot Interaction
}

\author{Yuan Gao$^{1*}$, Elena Sibirtseva$^{2*}$, Ginevra Castellano$^{1}$ and Danica Kragic$^{2}$
\thanks{*Authors contributed equally}
\thanks{$^{1}$Yuan Gao and
Ginevra Castellano are with the Department of Information and Technology, Uppsala University, Sweden. {\{\tt\small alex.yuan.gao, ginevra.castellano\}@it.uu.se}}%
\thanks{$^{2}$Elena Sibirtseva and
Danica Kragic are with the Robotics, Perception and Learning Lab, EECS at KTH Royal Institute of Technology, Stockholm, Sweden. {\{\tt\small elenasi, dani\}@kth.se}}%
}

\begin{document}

\maketitle
\thispagestyle{empty}
\pagestyle{empty}

\begin{abstract}
In socially assistive robotics, an important research area is the development of adaptation techniques and their effect on human-robot interaction. We present a  meta-learning based policy gradient method for addressing the problem of adaptation in human-robot interaction and also investigate its role as a mechanism for trust modelling.
By building an escape room scenario in mixed reality with a robot, we test our hypothesis that bi-directional trust can be influenced by different adaptation algorithms. We found that our proposed model increased the perceived trustworthiness of the robot and influenced the dynamics of gaining human's trust. Additionally, participants evaluated that the robot perceived them as more trustworthy during the interactions with the meta-learning based adaptation compared to the previously studied statistical adaptation model.


\end{abstract}


\section{Introduction}

In order to navigate in natural, dynamic environments, populated with humans, robots need to learn how to act, collaborate and adapt to different situations. For example, an assistive robot, that supports the elderly, needs to understand the person's need, performs basic object manipulation tasks, provides emotional support when needed, and more. Like humans, robots need the ability to adapt their behaviour and learn new ones through interaction with humans and other robots~\cite{rahwan2019machine}. These abilities are integral to achieving smooth human-robot interaction (HRI) for socially assistive robots.

Despite the fact that there exists an extensive body of work on application-driven adaptation in HRI, 
the fast adaptation that is grounded in realistic perception remains a challenge \cite{sheridan2016human}.
Recent developments have explored different aspects of human-robot interaction including physical human-robot interaction \cite{ghadirzadeh2016sensorimotor}, automatic reasoning~\cite{clark2018deep} and affective human-robot interaction~\cite{yuan2018when}.
All these methods are successful in their own field, but view the social HRI process from a narrower perspective. Also, they lack the flexibility to be extended with more complex, learning-based perceptual models. However, more complex models require more training data, and collecting data from HRI experiments is rather complicated.

In our work, we aim at developing a general approach for fast adaptation in HRI using a neural network-based policy gradient method. Similarly to earlier work~\cite{leite2014empathic}, we model the interactions as adversarial multi-armed bandit (MAB) problems~\cite{auer2002finite}. We address the sample inefficiency problem of policy gradient method using a meta-learning algorithm, called model-agnostic meta-learning (MAML)~\cite{finn2017model}. In Section \ref{sec:Method}, we provide a formal description of our model. 

One of the fundamental purposes of meta-learning is to give the agent the ability to draw conclusions on similar but unknown problems based on prior knowledge. This is very similar to how trust can shape people's behaviour when they need to interact with unknown humans for the first time. In psychological studies, it is shown that trust is a result of dynamic interactions~\cite{mayer1995integrative}. Successful interactions will lead to feelings of security, trust and optimism, whilst failed interactions will bring unsecured feelings or mistrust. We suggest that the pre-training phase of the meta-learning could be interpreted as gaining \emph{trust} towards the agent that the robot is going to interact with.

\begin{figure}[!t]
\includegraphics[width=0.46\textwidth]{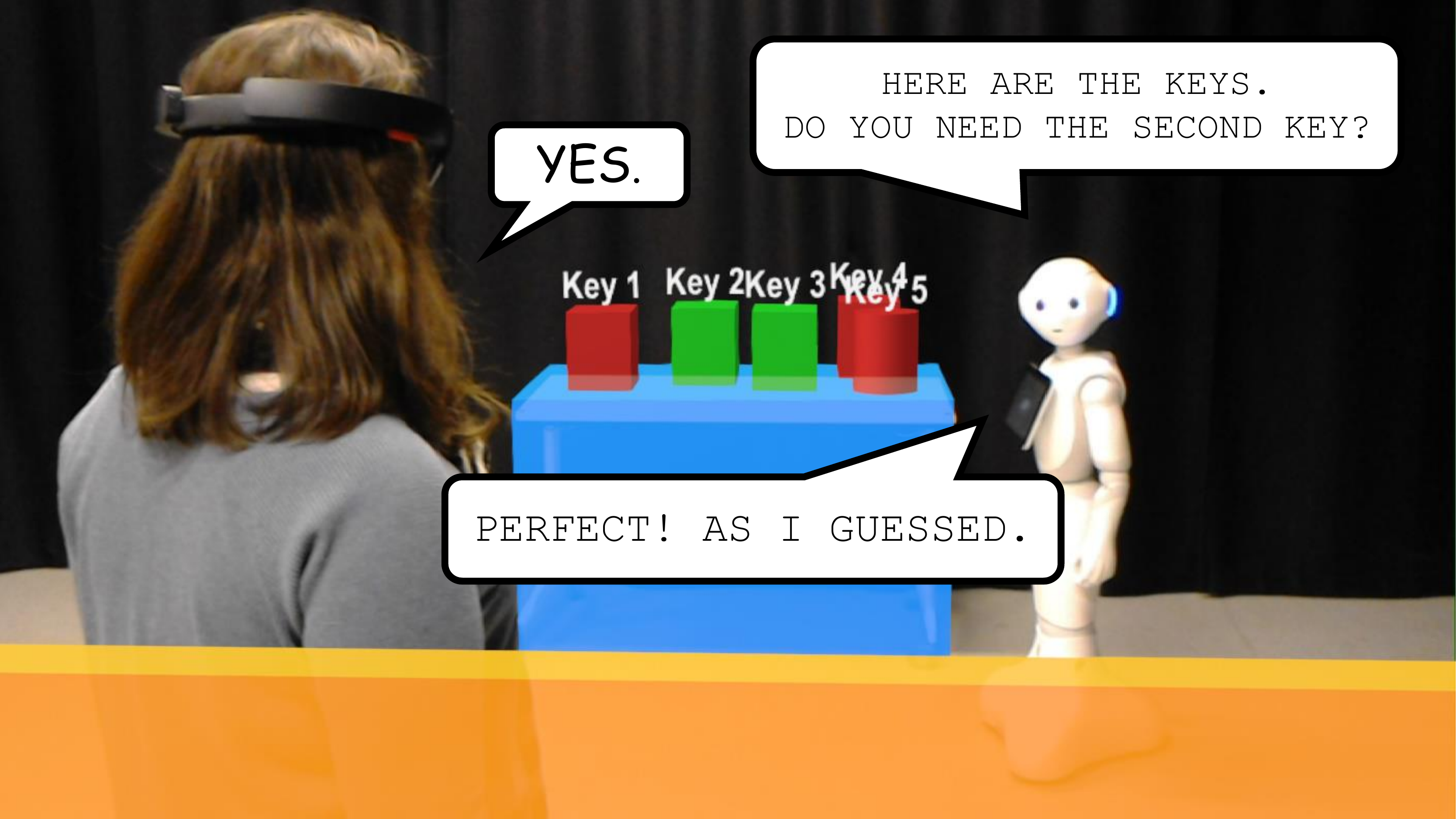}
\caption{This figure shows one of the interaction processes from the third-person point of view. The participant wears a Mixed Reality Headset, through which she can observe virtual objects (keys, blue table and orange walls of the escape room maze) augmented into the real world. 
}\label{hololens}
\end{figure}
In order to examine how our proposed model influences the perceived human's bi-directional trust, we evaluate it in a human-robot interaction scenario. In this scenario, a participant has to collaborate with a robot to escape a room, created with mixed reality (MR). From speech and human's 3D position in space, the robot learns to adapt to the needs of the participants. Our scenario is discussed in more details in Section \ref{sec:Experiment}. We used MR to create a dynamic environment that reacts to the person's and robot's behaviour, since MR presents additional benefits of flexibility and the speed of prototyping. Moreover, our previous studies \cite{sibirtseva2018comparison} demonstrated that MR does not affect the task performance in human-robot interaction compared to other traditional media, even taking into account the novelty effect of the new technology.

We investigate the effect of our adaptation model on the perceived bi-directional trust. Our main hypothesis is that, due to faster adaptation, our model increases the participant's perception of the robot's trustworthiness and how much, in their opinion, the robot trust them. We demonstrate the results of the human study in Section \ref{sec:Results} and discuss the future work in Section \ref{sec:Discussion}. 

The three main contributions of this paper are:



\begin{enumerate}
    \item We propose a policy gradient method based on meta-learning, which can be pre-trained in simulated auxiliary environments and adapt fast in a real-world HRI scenario.
    \item We propose that the meta-learning process can be viewed as a part of trust modelling, based on the psychological and sociological trust formalization in human-human interaction.  
    \item We performed a human-study to investigate the effect of our model on the subjective measures and found that our model increased the perceived bi-directional trust.
\end{enumerate}





\section{Related Work} 




In our work, to achieve fast adaptation, we utilize human feedback and use it under the framework of reinforcement learning (RL). 
RL has been used since the early days of research in social HRI. One of the early works was conducted by Bozinovski et al.~\cite{bozinovski1996emotion}, who considered the concept of emotion in its learning and behavioural scheme. Later, several researchers in HRI investigated the effect of RL algorithms like Exp3~\cite{leite2014empathic, yuan2018when, ahmad2018emotion} or Q-learning~\cite{mataric1997learning, tsiakas2018task}. With the development of deep learning~\cite{lecun2015deep}, several methods were proposed to understand different modalities in HRI, for example, ResNet~\cite{he2016deep} for image processing and Transformer~\cite{vaswani2017attention} based solutions for text processing. Naturally, the same trend can be observed regarding the problem of adaptation in HRI. One of the pioneer works was conducted by Qureshi in 2017~\cite{qureshi2016robot} where a Deep Q-Network~\cite{mnih2015human} was used to learn a mapping from visual input to one of the several predefined actions for greeting people. 

However, for deep RL methods, one of the drawbacks is that the algorithms need a lot of training data to converge~\cite{mnih2015human}. This makes deep learning methods' applicability in HRI limited, even for the most basic adaptation problem.
Thus, the dilemma is clear: on the one hand, if statistical methods are used, the robot may not be able to capture the details of the interactive process~\cite{yuan2018when}. On the other hand, if deep learning methods are used, a lot of training data is needed to optimize the algorithms~\cite{mnih2015human}. One potential solution to the dilemma may be to use  meta-learning. Using meta-learning for pre-training may address the problem of limited training data~\cite{duan2016rl,finn2017model}. Examples of this have been shown for image recognition~\cite{zoph2018learning} and imitation learning~\cite{duan2017one}. However, to the best of our knowledge, there is no research on applying meta-learning in HRI.  


Trust as a social phenomenon has been widely studied in psychology, sociology, and economics, however, the definition of trust is not commonly agreed upon across the disciplines. 
It was shown in numerous studies that trust is essential for successful human-human interactions \cite{mcallister1995}. In regards to relationships with robots, trust plays a major role in human's willingness to accept information from a robot \cite{hancock2011meta} and to cooperate \cite{freedy2007measurement}. 
Modelling trust can help us to build more intelligent socially assistive robots.

According to Marsh's formalization, trust consists of three components, namely, basic, general and situational trust of an agent in another agent  \cite{marsh1994formalising}. The basic trust is the value derived and updated from the previous experience, and it helps the agent to make decisions about the unknown agents in future situations. The general trust is agent-specific, representing a bias towards another particular agent, while the situational trust is dependent on external conditions. In our case, we focus on modelling the basic trust, which can be viewed as a general disposition of the robot to be more trustworthy towards a human during an interaction. As a consequence, this can increase the speed of adaptation in a particular task. From the sociological standpoint, the convergence of the meta-reinforcement learning process can be considered as a part of trust modelling.


\subsection{Trust evaluation in HRI}

There is no commonly used procedure to measure perceived trust in HRI. As stated in Hancock's overview of the field, one of the most influential factors on perceived trust is a robot's performance, its reliability and understandability \cite{hancock2011meta}. Another study showed that human-related subjective measures, such as personality traits and level of expertise, also have an effect on how people perceive the robot's trustworthiness \cite{salem2015would}. Moreover, in human-computer interaction literature, cooperation is defined as a "behavioural outcome of trust" \cite{wilson2006all}. We combine all previously mentioned metrics together to measure perceived trust during our human study. In the majority of works, trust was measured after the interaction (e.g. \cite{schaefer2013perception} and \cite{aroyo2018trust}). However, trust is dynamic in nature and can be influenced by many different factors throughout an interaction \cite{BLOMQVIST1997271}. To capture the changing dynamics of trust, it is advised to measure it multiple times during an interaction \cite{mcallister1995, schaefer2016measuring}.
 

Another thing to consider in trust evaluation is the HRI scenario. A common scenario to evaluate trust involves economic games, where a participant is asked to gamble for a monetary gain \cite{lee2013computationally}. Some works, on the other hand, present a more general task, where a robot also asks participants to perform unusual requests \cite{salem2015would}. In both cases, only one-directional trust towards the robot is measured. However, to our knowledge, there is no study that measures the bi-directional perceived trust in HRI. 
In this work, we develop an escape room scenario that allows variety in robot's and human's behaviours and is suitable for testing bi-directional perceived trust.
The escape room scenario has been found to be helpful for investigating human's behaviour in diverse collaborative tasks \cite{pan2017collaboration}. Recently, a similar scenario has been used to study the effects of robot's failure in human-robot collaboration~\cite{van2019take}. In an escape room scenario, players are locked in a room and they need to solve a series of puzzles in order to escape it under a time constraint. The main advantage of this scenario is its flexibility, which means that specific puzzles can be easily added to study different behaviours. 

\section{Interaction Modelling}\label{sec:Method}
In this section, we describe our approach to model the human-robot interaction process. In Section~\ref{method:preliminaries}, we first introduce the general RL problem mathematically. Then in Section~\ref{method:our_model}, we describe the details of our model implementation.
\subsection{Preliminaries}\label{method:preliminaries}
We consider a general interaction process, modelled using $\mathbf{s}_t$ and $\mathbf{a}_t$ as the state and action of the robot agent at time $t$ during the interaction. The interaction could be viewed as maximization of the expected cumulative reward $E_{\tau\sim\pi}[\mathcal{R}(\tau)]$ over trajectories
\begin{align}
\tau = \{\mathbf{s}_1,\mathbf{a}_1,\dots, \mathbf{s}_T, \mathbf{a}_T\},
\end{align}
where $T$ is the final time step of the interaction, $\mathcal{R}(\tau)=\sum_{t=1}^T\mathcal{R}(\mathbf{s}_t, \mathbf{a}_t)$ is the cumulative reward over $\tau$. The expectation is under a distribution, 
 \begin{align}p(\tau) = p(\mathbf{s}_1)\prod_{t=1}^{T}p(\mathbf{s}_{t+1}|\mathbf{s}_t,\mathbf{a}_t)p(\mathbf{a}_t|\mathbf{s}_t),
 \end{align}
 where $\pi(\mathbf{s}_t) = p(\mathbf{a}_t|\mathbf{s}_t)$ is the policy we would like to train and $p(\mathbf{s}_{t+1}|\mathbf{s}_t,\mathbf{a}_t)$ is the forward model determined by the interaction dynamics. We now define a meta-learning technique $M \equiv \{\zeta_p, \zeta_r\}$ consisting of a pre-training method $\zeta_p$ and a refinement method $\zeta_r$. Here, $\zeta_p$ takes policy $\pi$ and auxiliary environments $\mathcal{E}_x$ to receive a meta-policy
 
\begin{align}
\pi_{meta} = \zeta_p(\mathcal{E}_x, \pi).
\end{align}
Intuitively, the meta-policy is a policy that learned prior knowledge about the tasks it is going to solve. For more detailed explanation, interested readers refer to this survey paper~\cite{vanschoren2018meta}.
 
We then consider a task-specific environment $\mathcal{E}_t$ for policy refinement. The final outcome of the system, that we would like to have, is the sampled trajectories $\tau^\star_i$ from all possible trajectories generated under the optimized policy $\pi^\star$ using $\zeta_r$. Mathematically, $\tau^\star_i$ is defined as $\tau^\star_i \sim \pi^\star$, where $\pi^\star = \zeta_t(\mathcal{E}_t, \pi_{meta})$.

\subsection{Proposed model}\label{method:our_model}
In order to model the interaction in our scenario, we loosely follow the assumption that a human-like robot should have the tripartite mental activities, namely conation $\mathcal{T}$, cognition $\mathcal{G}$ and affection $\mathcal{A}$. The assumption is inspired by Hilgard's tripartite classification of mental activities of human personality and intelligence in modern behaviour psychology  \cite{hilgard1980trilogy}. Here we define the interactive space $\mathcal{C}$ as $\mathcal{C} \equiv \{\mathcal{T}, \mathcal{G}, \mathcal{A}\}$. For our particular escape room scenario, each instance of the mental functionality, $\mathcal{E}^{c \in \{\mathcal{T}, \mathcal{G}, \mathcal{A}\}}$, is then modelled and implemented as an environment of adversarial MAB problem. All of the three instances operate independently throughout the interaction process. We define different meta-learning processes $M^{c \in \{\mathcal{T}, \mathcal{G}, \mathcal{A}\}}$ for different functionality in $\mathcal{C}$. Each meta-learning strategy contains $\{\zeta_p^{c}, \zeta_r^{c}\}$.

During the interaction, the robot needs to optimize its all policies $(\pi^c)^*$ to learn the most preferred action for each MAB environment. In order to keep the generality of the concept of trust, we also assume the observational states of each category $\mathbf{s}^c$ to be fixed for each category $c$.

Our methods involve two training processes. Firstly, we model the human feedback of each action of MAB as a Gaussian distributions $r^c \sim \mathcal{N}(\mu^c,(\sigma^c)^{2})$ for all the auxiliary environments. This modelling is based on the fact that signal of emotion recognition normally follows Gaussian distributions~\cite{kragel2016decoding}.  Simultaneously, we use $\zeta_p^{c}$ to train initial random $\pi^c$ in order to get a meta policy $\pi_{meta}^c$. $\pi_{meta}^c$ learns the inner structure of the problem which makes the adaptation during the interactive session much faster and data-efficient. Finally, we conduct human experiments and study different subjective measures along with interaction. Mathematically, the training steps can be summarized as follows:
\begin{align}
    \mathcal{R}^c(\mathbf{s}, \mathbf{a}) &\equiv r^c \sim \mathcal{N}(\mu_\mathbf{a}^c,(\sigma_\mathbf{a}^c)^{2})\quad\forall \mathcal{E}_x^c\\
    \pi_{meta}^c &= \zeta_p^{c}(\mathcal{E}_x^c, \pi^c)\\
    (\pi^c)^* &= \zeta_p^{r}(\mathcal{E}_t^c, \pi_{meta}^c),
\end{align}
Where $c\in\{\mathcal{T}, \mathcal{G}, \mathcal{A}\}$, $\mu_\mathbf{a}^c$ and $\sigma_\mathbf{a}^c$ are mean and standard deviation associated with action distribution $\mathbf{a}$, $\mathcal{E}_x^c$ are auxiliary environments for pre-training and $\mathcal{E}_t^c$ are real interactive environments. We use MAML ~\cite{finn2017model} as the meta-learning algorithms $M^c$ and trust region policy optimization (TRPO)~\cite{schulman2015trust} as optimization algorithm for all policies. Interested readers can refer to Appendix~\ref{ramdom_meta} to see the differences in convergence between meta policy and randomly initialized policy for MAB problems with different number of actions using a neural network based policy.

\textbf{Implementation details:} We used a PyTorch based MAML implementation\footnote{https://github.com/tristandeleu/pytorch-maml-rl} and built a customized MAB environment using OpenAI Gym~\cite{brockman2016openai}. For each instance of mental functionality, an auxiliary environment is used as the pre-training method. For all instances, the same policy network structure is used, namely, one hidden layer with 100 neurons. MR engine Unity is used to animate the changes in the Hololens environment\footnote{https://unity.com/}. The communication between Hololens and the algorithm in the policy refinement step is carried out via a UDP based communication method.

\section{Methodology}\label{sec:Experiment}




\subsection{Scenario}

\begin{figure}[h]
\includegraphics[width=0.46\textwidth]{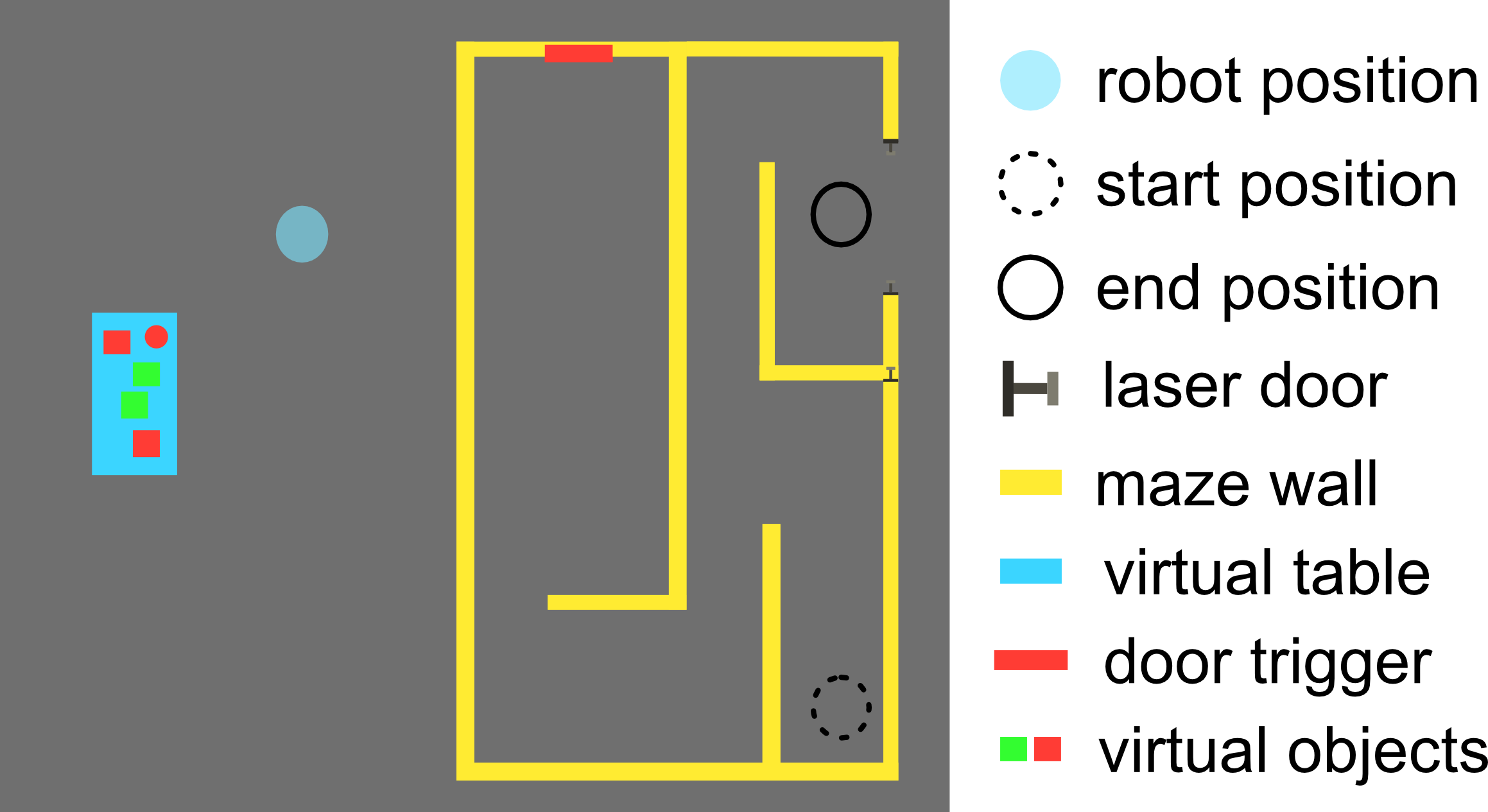}
\caption{This figure shows the setup of our study. During a run, the participant starts from the dotted line circle to approach the red button trigger. Then it goes to the solid line circle to go out of the room. During the run, three instances of mental activities will each be triggered once. }\label{setup}
\end{figure}

We built an escape room scenario to conduct a between-subjects study, in which participants interacted with a Pepper robot\footnote{https://www.softbankrobotics.com/emea/en/pepper}. The escape room was created in augmented reality and participants were required to wear a Mixed Reality headset HoloLens\footnote{https://www.microsoft.com/en-us/hololens} to see walls of the virtual maze, triggers, keys, and the exit door (see Fig~\ref{setup}). The interaction consists of three parts: an instance of conation, an instance of affection, and an instance of cognition. Each instance is triggered by the participant's position in the virtual maze, recorded from the HoloLens. For each instance, the robot chooses one out of four actions, according to a probability distribution provided by the algorithm. Here, an action is implemented as a verbal question (i.e. \textit{``Did you come here to bring me something?"}, see Table \ref{tab:questions}). After the participant answers, the robot updates the probability associated with the previous question and gives feedback accordingly (i.e. for the low probability ``\textit{I do not believe it, but fine.}", see Table \ref{tab:replies}). After completing all the interaction steps, the participant can escape the room and a run is over. 

For the control group, a statistical MAB algorithms Exp3 (\textbf{C1}), implemented as in the previous studies, was used. The experimental group interacted under our proposed model, a policy gradient based solution for MAB problem, together with meta-learning (\textbf{C2}).

\begin{table}[t]
\centering
\resizebox{\columnwidth}{!}{\begin{tabular}{ll}
\hline
Mental activity & Examples of implementation \\ \hline
Conation                 & \begin{tabular}[t]{@{}l@{}}``Do you want to escape the room?''\\ ``Do you come here to stay with me?"\\ ``Did you come here to bring me something?"\\ ``Did you come here to look for your friends?"\end{tabular} \\
Affection                & \begin{tabular}[t]{@{}l@{}}``Be careful with the walls. You should avoid them."\\ ``I can help you to do whatever you want to do here."\\ ``Hey, be relaxed, no matter what you do,\\you have no fault in this game."\\ ``Are you worried? Don't worry! I am here with you."\end{tabular} \\
Cognition                & \begin{tabular}[t]{@{}l@{}}``Here are the keys. Do you need the first key?"\\ ``Here are the keys. Do you need the second key?"\\ ``Here are the keys. Do you need the third key?"\\ ``Here are the keys. Do you need the fourth key?"\end{tabular}\\ \hline
\end{tabular}}
\caption{Examples of questions the robot asks during an interaction based on the mental activity classification \cite{hilgard1980trilogy}.}\label{tab:questions}
\end{table}


\subsection{Procedure}
 
Upon arrival, participants received a brief description of the experiment. Then they were asked to sign a consent form and fill in a pre-study questionnaire about their demographic background, prior experience with robots and technology, a personality assessment \cite{gosling2003very}, and negative attitude towards robots  \cite{syrdal2009negative}. In order for the algorithms to converge, the participant's behaviour has to be consistent during several runs. Thus, the participants are asked to act according to the following rules during the interaction:

\begin{itemize}
    \item Your goal is to escape the room;
    \item You need to get further information regarding the walls of the maze;
    \item You need the second key;
    \item You can answer only ``yes" or ``no";
\end{itemize}

The first attempt at escaping the room is treated as a test run to familiarize the participant with the setup and is not taken into account in an adaptation algorithm. Each participant has to go through four sessions, while each session consists of three full runs, during which the robot gradually adapts to the participant's requests. Overall, each participant goes through twelve iterations of algorithm adaptation. After each session, the participant fills in a short questionnaire evaluating perceived bi-directional trust. By the end of the fourth session, the participant fills in a questionnaire regarding the overall experience and the quality of interaction.


\subsection{Hypotheses}

By comparing the effects of the two conditions we aim to test whether the \textit{overall perceived bi-directional trust} is affected by different adaptation algorithms. We propose to define \textit{bi-directional trust} as how trustworthy the participant perceives the robot and how much, in their opinion, the robot trusts them in return. Moreover, we hypothesize that the \textit{dynamics} of how the bi-directional trust changes throughout the interaction sessions vary in two conditions. More formally, 

\begin{itemize}
    \item \textbf{H1}: Perceived trust of a participant towards the robot is \textit{higher} in the \textbf{C2}.
    \item \textbf{H2}: The dynamics of perceived trust of a participant towards the robot differs in two conditions.
    \item \textbf{H3}: Perceived trust of a robot towards the participant is \textit{higher} in \textbf{C2}.
    \item \textbf{H4}: The dynamics of perceived trust of a robot towards the participant differs in two conditions.
\end{itemize}

Based on the simulation results (see Fig.~\ref{simu_exp3_meta}), we assume that our proposed model (condition \textbf{C2}) ,with faster adaptation and its specific adaptation dynamic, will influence participant's perception of bi-directional trust in a positive way. Additionally, we expect that slower adaptation (condition \textbf{C1}) will cause fewer changes in subjective measures between four sessions, in comparison to the meta-learning based approach. 

\begin{table}[t]
\centering
\begin{tabular}{ll}
\hline
Confidence level & Examples of robot's replies         \\ \hline
Low ($p<0.5$)             & ``I do not believe it but fine."     \\
Medium-low ($p<0.65$)      & ``Is that really so? I am not sure." \\
Medium ($p<0.8$)          & ``I understand now."                 \\
High ($p\geq$0.8)             & ``Awesome, I knew you would say so." \\ \hline
\end{tabular}
\caption{Examples of robot's replies to participants' answers based on the confidence level, where $p$ denotes probability associated with the action.}\label{tab:replies}
\end{table}

     

\subsection{Measures}

\subsubsection{Objective measures}

In order to analyze the performance of the algorithms in our setup, we tested our method using simulated data. Like what we described in the previous section~\ref{method:our_model}, we assume user's feedback is a continuous signal distributed with Gaussian function with mean $\mu=1$ and variance $\sigma^2=0.1$. Based on this simulated feedback, we compare the adaptation speed between the Exp3 algorithms in the control group and the meta-policy in the experimental group. In this simulation, we choose the number of actions in MAB settings to be four, similar to previous studies~\cite{yuan2018when,leite2014empathic}.

\subsubsection{Subjective measures}
Bi-directional perceived trust is measured from the participant's point of view, how trustworthy they perceive the robot and how much they think the robot trusts them in return. We evaluate the bi-directional perceived trust by following Salem et al.'s work on trust evaluation in HRI \cite{salem2015would}. Single items were extracted and adjusted to suit our scenario. 

Subjective measures were collected in a form of a questionnaire, given to the participants after each session during the interaction. The participants were asked to evaluate their answers on the 5 point Likert scale (1 = ``\textit{strongly disagree}", 5 = ``\textit{strongly agree}"). 

We selected single modified items from \cite{salem2015would}: ``I perceive the robot as trustworthy", ``The robot perceives me as trustworthy". Single modified item was added from the ``Propensity to Trust survey" \cite{evans2008survey}: ``The robot anticipates my needs". 
We further examined participant's perception of the robot's trust towards them with counteractive items: ``The robot believed my answers", ``The robot questioned my answers". 















\subsection{Participants}

A total of 24 subjects (11 female, 13 male), with ages ranging between 22 and 48 ($M=27.04$, $SD=5.42$), were recruited for this experiment. On a Likert scale from 1 to 5 (with 1 representing very little and 5 - very much), participants were found to have moderate experience interacting with robots ($M=3.04$, $SD=1.338$), negligible familiarity with Virtual or Augmented reality ($M=2.32$, $SD=1.376$), and major skills regarding digital technologies ($M=4.76$, $SD=.436$). Participants were randomly assigned to one of the two conditions, resulting in two groups of 12 subjects each. 

\section{Results}\label{sec:Results}

\subsection{Objective Measures}
\begin{figure}[h]
\includegraphics[width=0.46\textwidth]{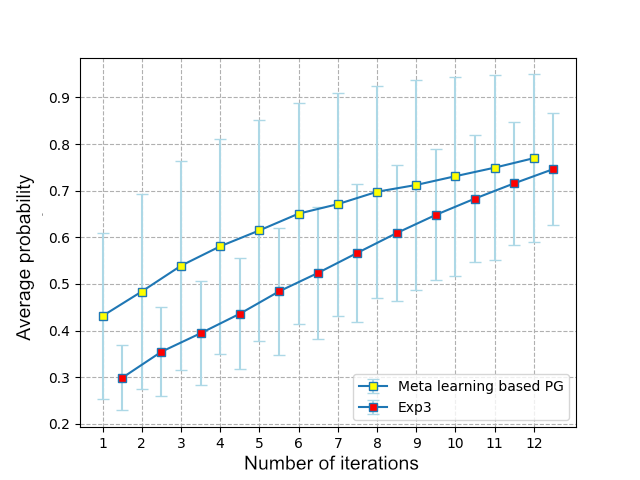}
\caption{This figure shows the simulated adaptation results of the escape room scenario for two different algorithms after each interaction. The results of Exp3 algorithms is shifted to the right in order to show the difference clearly.}\label{simu_exp3_meta}
\end{figure}

 Fig.~\ref{simu_exp3_meta} shows the comparison between the overall adaptation speed between the control group and the experimental group over all the instances using simulated feedbacks. The x-axis indicates the average probability of all the right answers over all the instances throughout the four sessions. The y-axis shows the number of iterations that the algorithms has optimized. The result shows that our method has on average higher adaptation speed.

\subsection{Subjective Measures}
The bi-directional perceived trust was analyzed using a mixed design repeated measures one-way ANOVA at significance level $\alpha=.05$. The two conditions were viewed as between-subjects factors $(N_1=12, N_2=12)$, while sessions numbers corresponded to within-subjects factor on the dependent variables of perceived trust towards the robot and perceived robot's trust towards the participant. The overall measurements were compared based on the results of the tests of the between-subjects effects. Neither of the two measures violated the sphericity assumption. Dynamics analysis was carried out on the basis of the interactions' statistical significance.

\subsubsection{Perceived trust towards the robot} 

\begin{figure}[t]
\includegraphics[width=0.46\textwidth]{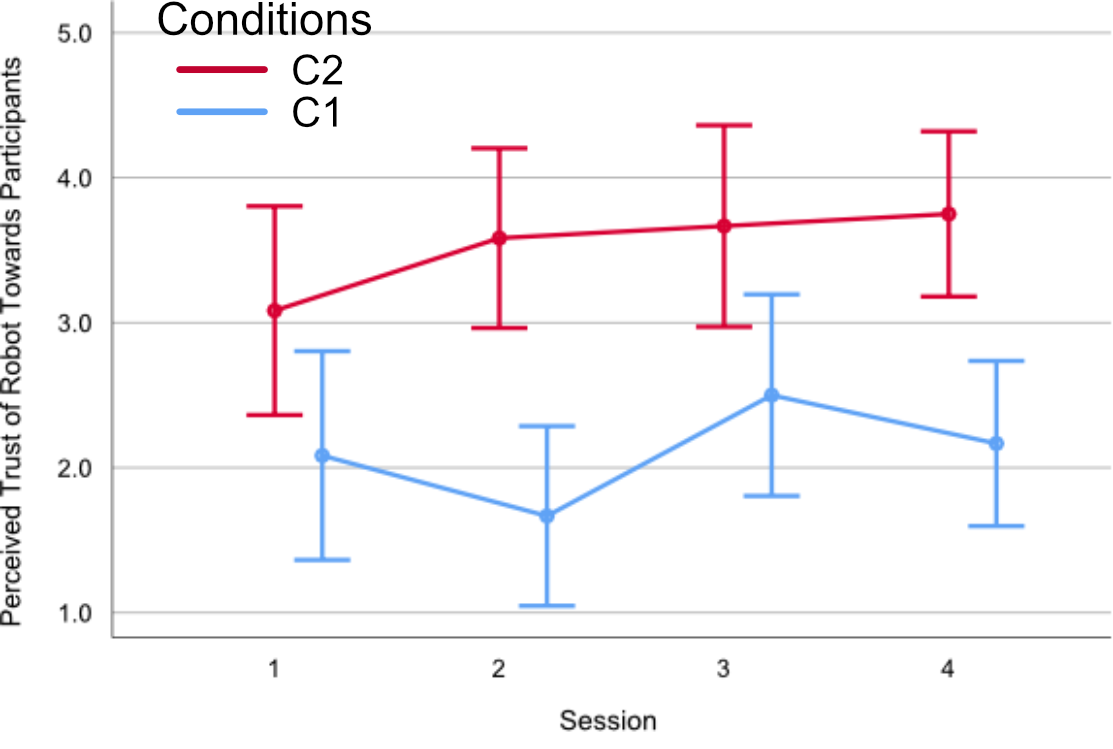}
\caption{The average value (5-point Likert scale) of the participant's perceived trust towards a robot by session number and condition with 95\% CI errors. The condition \textbf{C1} is shifted to the right to show the difference clearly.}\label{results:robot-trust}
\end{figure}

For the condition \textbf{C1} (see Fig.~\ref{results:robot-trust}), the first session resulted into moderate scores ($M=$ 3.03, $SD=$ 1.24), then the perceived trust towards the robot dropped in the second session ($M=$ 2.42, $SD=$ .996). After that, it stabilized during the third ($M=$ 2.83, $SD=$ 1.267) and the fourth ($M=$ 2.67, $SD=$ .888) sessions. In the condition \textbf{C2}, at the beginning participants evaluated their perceived trust towards the robot at a similar moderate score as the C1 ($M=$ 2.92, $SD=$ .793). In the consecutive sessions, we can observe a gradual increase of the scores, specifically after the second session ($M=$ 3.17, $SD=$ .835), the third ($M=$ 3.75, $SD=$ .754), and finally the fourth ($M=$ 3.92, $SD=$ .515). 

We found a statistically  significant effect of condition on participant's perception of the robot's trustworthiness, $F(1,22)= 4.794, p < .05, \eta^{2}=  .179$, such that the average score of the \textbf{C1} ($M=$ 2.75, $SD=$ .222) is lower than the \textbf{C2} condition score ($M=$ 3.437, $SD=$ .222). Thereby, \textbf{H1} \textit{was supported} and the participants perceived the robot's trustworthiness higher in the meta-learning based adaptation.
A statistically significant effect of the session number was found, $F(3,66)= 1.427, p < .02, \eta^{2}= .142 $, regardless of condition.

There was a statistically significant effect of the session number and condition interaction, $F(3, 66)= 5.629, p < .002, \eta^{2}=  .204$. This means that the dynamics of how the participants gain trust in the robot is significantly different in two conditions and \textbf{H2} \textit{was supported}. 


\begin{figure}[t]
\includegraphics[width=0.46\textwidth]{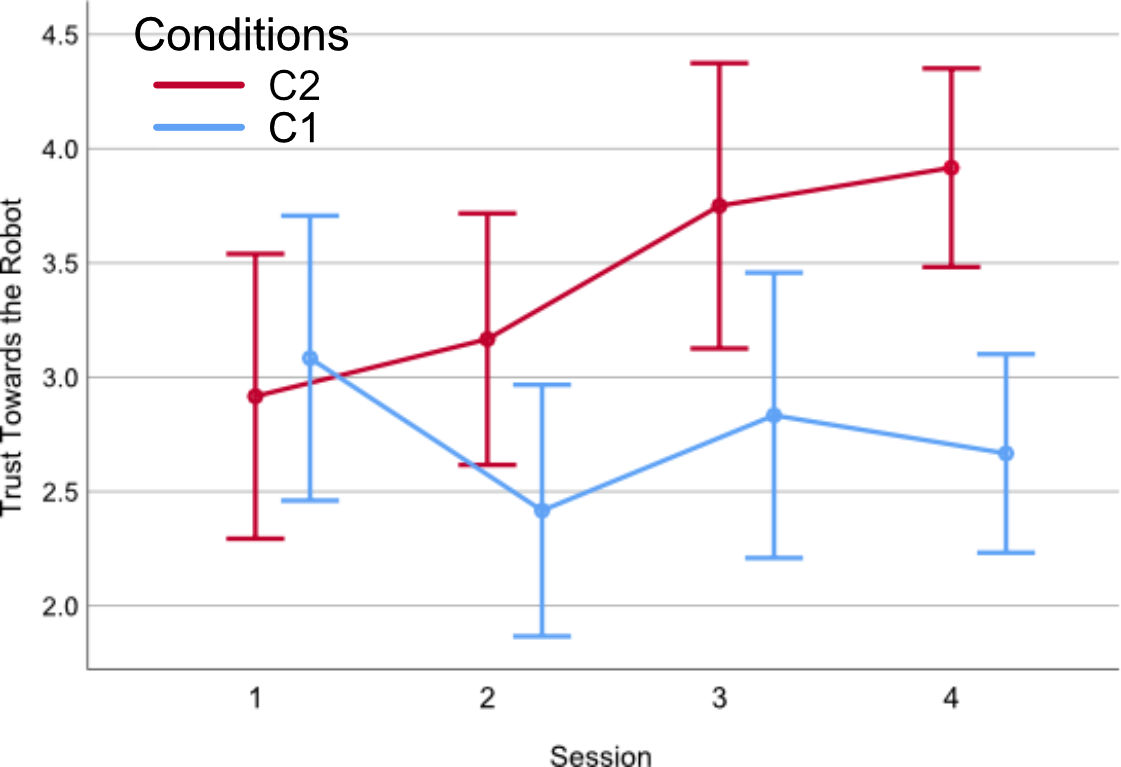}
\caption{The average value (5-point Likert scale) of the robot's perceived trust towards a participant from a robot by session number and condition with 95\% CI errors. The condition \textbf{C1} is shifted to the right to show the difference clearly. }\label{results:human-trust}
\end{figure}

\subsubsection{Perceived robot's trust towards the participant} 

From Fig.~\ref{results:human-trust}, we can observe that in the beginning, the participants in the condition \textbf{C1} evaluated moderately-low ($M=$ 2.08, $SD=$ .996) on how trustworthy the robot perceives them. In the second session, the average score decreased ($M=$ 1.67, $SD=$ .778) and then increased in the third session ($M=$ 2.5, $SD=$ 1.168). The scores in the final session decreased ($M=$ 2.17, $SD=$ .937). The results of \textbf{C1} show fluctuations between sessions, but also have a positive trend overall. In comparison to the condition \textbf{C1}, the first session in condition \textbf{C2} has higher average score ($M=$ 3.08, $SD=$ 1.379), which increased after the second session ($M=$ 3.58, $SD=$ 1.24), and flattened out afterwords in the both third ($M=$ 3.67, $SD=$ 1.155) and fourth ($M=$ 3.75, $SD=$ .965) sessions.

A statistically significant effect of condition was found on perceived trust towards the participant, $F(1,22)= 16.44, p < .001, \eta^{2}= .428$, such that the \textbf{C2} has higher average score of perceived trust towards the participant ($M=$ 3.521, $SD=$ .247) compared to the \textbf{C1} ($M=$ 2.104, $SD=$ .247). This denotes that the participants perceived the robot as more trustworthy towards them in the \textbf{C2} and \textbf{H3} \textit{was supported}. 

Even though the overall robot's trust towards the participant depends on condition, the differences in dynamics of how the perceived robot's trust towards the participant changes in two conditions are not statistically significant, $F(3,66)= 1.673, p = .181, \eta^{2}= .071$, \textbf{H4} \textit{was rejected}. Additionally, no statistical significance of session number was found, $F(3,66)= 2.374, p = .078, \eta^{2}= .097$.

\section{Discussion}\label{sec:Discussion}

Despite \textbf{H3} \textit{being  supported} and participants perceiving the robot as more trusting towards them in the \textbf{C1}, the dynamics of the perceived robot's trust towards the participant was not significantly different between groups. As a consequence, \textbf{H4} \textit{was not supported}. This can be explained by the meta-learning based adaptation algorithm converging on at least one of the actions during the first session. In other words, during the first session the robot might have gone from explicitly saying ``I don't believe you" to a more neutral ``I understand". Thus, the results show a significant difference between the first sessions in two conditions, but for the subsequent sessions, the perceived robot's trust towards the participant fluctuates within the 95\% confidence interval. We can hypothesize that we observe this due to too fast adaptation with the meta-learning pre-trained policy gradient algorithm and too slow adaptation with Exp3.

In contrast, the convergence rate of the algorithms did not have the same effect on how trustworthy the participants perceived the robot. The estimated marginal means in the second condition were significantly higher, meaning that the participants found the robot more trustworthy in the second group. Thus, we can conclude that \textbf{H1} was supported. However, we also found a significant effect of interaction, which can be interpreted as a difference in trends for two conditions and supports \textbf{H2}. A noteworthy distinction between the dynamics of  the robot's trust towards the participant and the trust towards the robot appears in the first session. 

The discrepancies of how two algorithms influenced the dynamics of perceived bi-directional trust can be explained by the explicit nature of how the robot expressed its trust towards the participants, while human trust is a complex social construct. Further investigation is required in order to examine this phenomenon. 

Compared to \cite{yuan2018when}, we established an alternative method to model the interactive adaptive process and show that our model is able to learn a prior from modelled auxiliary environments. This helps the algorithm to deal with the real-time requirement in human-robot interaction.



We think this work could be extended in two directions, both from the algorithmic and the human study point of view. From the algorithmic perspective,
we want to incorporate deep learning perceptual modules for processing multimodal human input to enrich the interaction process.
Second, trust modelling can be expanded to include implementation of general and situational trust, as well as reaction to trust violation, according to Marsh's formalization \cite{marsh1994formalising}. Finally, the developed system can be tested further for how different approaches of trust modelling influence the participants' perception of trust, robot's intelligence, and quality of interaction. 

We hypothesize that the proposed models can help us to achieve a richer socially interactive process for real-time HRI. 

\section{Conclusion and Future Work}
In this work, we proposed to use a meta-learning based policy gradient method for addressing the problem of fast adaptation in HRI. Compared to the statistical model, it can be pre-trained in auxiliary environments and then adapted faster in the real HRI scenario.

We designed an escape room scenario in mixed reality to evaluate the proposed method and investigate its potential effects on the perceived bi-directional trust. Our results show that not only the algorithm adopted a higher learning rate after the meta-learning process but also has increased the participant's perception on how trustworthy the robot perceives them.


In the future, we will combine this modelling with different neural network-based perception modules and examine their influences on the interactive process. From the trust modelling side, we will investigate the possible cause of the differences in the dynamics of perceived bi-directional trust. Moreover, we will address trust violation and examine human's perception of different approaches to modelling it. 
\section*{Acknowledgement}
This work was supported by the COIN project (RIT15-0133) funded by the Swedish Foundation for Strategic Research and by the Swedish Research
Council (grant n. 2015-04378)

\bibliography{bibliography} 
\bibliographystyle{IEEEtranS}

\section*{Appendix}
\subsection{Comparison of meta policy and randomly initialized policy}\label{ramdom_meta}
In order to assess how meta-learning pre-training process influences the optimization of a neural network based solution for the multi-armed bandit problem, we compare the meta-policy with randomly initialized policy. The meta-policy is pre-trained with the simulated auxiliary environments. The feedback distribution in the auxiliary environments is modelled based on the previous research of modelling the perceived emotion using Gaussian class functions~\cite{zhang2017predicting}.

\makeatletter
\setlength{\@fptop}{0pt}
\makeatother

\begin{figure}[h]
\includegraphics[width=0.46\textwidth]{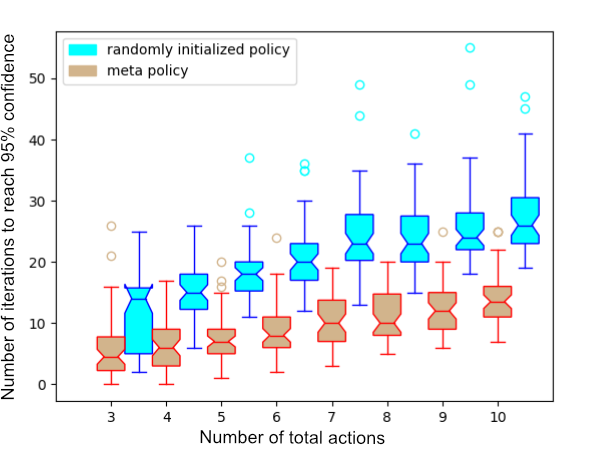}
\caption{This figure shows the relationship between the average number of samples needed to reach 95\% confidence for any action and the number of actions in MAB problems. The results of the randomly initialized policy are shifted to the right to show the differences clearly.}\label{simu_results}
\end{figure}

Fig.~\ref{simu_results} shows a comparison of the two classes of policies. We can observe that as the number of actions of MAB increases, the number of interactions needed to reach 95\% confidence for a particular action increases. However, the meta learned policy needs less number of iterations than a randomly initialized policy for all MAB problems with the different number of actions.

\end{document}